# Boosting R-CNN: Reweighting R-CNN Samples by RPN's Error for Underwater Object Detection

Pinhao Song[a], Pengteng Li[b], Linhui Dai[a], Tao Wang[a] and Zhan Chen[a]

[a]*Key Laboratory of Machine Perception, Shenzhen Graduate School, Peking University, Shenzhen, 518055, Guangdong, China*
[b]*Shenzhen University, Shenzhen, 518060, Guangdong, China*

## ABSTRACT

Complicated underwater environments bring new challenges to object detection, such as unbalanced light conditions, low contrast, occlusion, and mimicry of aquatic organisms. Under these circumstances, the objects captured by the underwater camera will become vague, and the generic detectors often fail on these vague objects. This work aims to solve the problem from two perspectives: uncertainty modeling and hard example mining. We propose a two-stage underwater detector named boosting R-CNN, which comprises three key components. First, a new region proposal network named RetinaRPN is proposed, which provides high-quality proposals and considers objectness and IoU prediction for uncertainty to model the object prior probability. Second, the probabilistic inference pipeline is introduced to combine the first-stage prior uncertainty and the second-stage classification score to model the final detection score. Finally, we propose a new hard example mining method named boosting reweighting. Specifically, when the region proposal network miscalculates the object prior probability for a sample, boosting reweighting will increase the classification loss of the sample in the R-CNN head during training, while reducing the loss of easy samples with accurately estimated priors. Thus, a robust detection head in the second stage can be obtained. During the inference stage, the R-CNN has the capability to rectify the error of the first stage to improve the performance. Comprehensive experiments on two underwater datasets and two generic object detection datasets demonstrate the effectiveness and robustness of our method. The link of code: https://github.com/mousecpn/Boosting-R-CNN

## 1. Introduction

Oceans account for 71% of the earth's total area and contain rich biological and mineral resources. Humans cast their eyes on ocean exploitation, for the resources on the land have been fully exploited, which means the research on the oceans is meaningful. Over the past few years, more and more researchers have considered applying underwater object detection (UOD) to autonomous underwater vehicles (AUVs) with visual systems to fulfill a series of underwater tasks such as marine organism capturing.

Generic Object Detection (GOD) has been researched for a long time and obtained abundant achievements. However, GOD is not perfectly suitable for underwater environments which bring new challenges to object detection (see Figure 1): (i) The images captured by the underwater visual system suffer from unbalanced light conditions and low contrast, which make the object hard to be distinguished from the background. (ii) The aquatic organisms tend to live together, which cause severe occlusion. (iii) The aquatic organisms are good at hiding themselves, which have the similar color with the background and make it hard for people to recognize them. Facing these new challenges, the boundaries between the objects and background and the boundaries between different objects will be vague, leading to the existence of the vague objects in underwater environments.

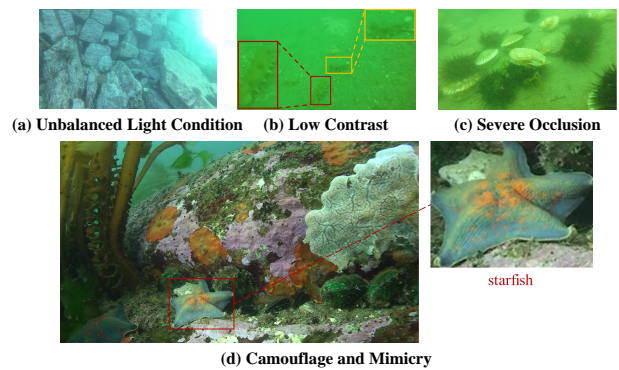

**Figure 1:** The challenges of underwater environments. (a) Complicated underwater terrains cause unbalanced light condition. (b) The low-contrast image makes the boundaries of two holothurian blurred. (c) The aquatic organisms tend to live together, causing occlusion. (d) The starfish has the similar color with the environment, which makes them difficult to spot.

Existing works on UOD typically apply data augmentation methods [19, 30, 47] and use a strong feature extractor [14, 57] to improve the performance. However, these methods suffer from problems listed as follows. **(i)** Previous underwater detectors receive the same supervision signal for all objects regardless of their vagueness. Thus, the classification score trained with simple cross entropy loss does not accurate reflect the vagueness of the objects, which would cause false over-confident predictions. However, accurately ranking the detection results is crucial for object detectors to achieve high performance. It is expected that the detectors assign low scores to the detection results containing vague objects and assign high scores to the results with clear objects. **(ii)** Previous underwater detectors are vulnerable to vague objects with blurring boundaries and similar color to the background. That is because the gradient of the easy samples will dominate the training of underwater detectors, which makes detectors difficult to learn the subtle differences between vague objects and underwater background.

✉ Pinhaosong@pku.edu.cn (P. Song); 2110276192@email.szu.edu.cn (P. Li); dailinhui@pku.edu.cn (L. Dai); taowang@pku.edu.cn (T. Wang); zhanchen\_cz@pku.edu.cn (Z. Chen)
ORCID(s):

Different from existing UOD methods, we address the above problems through uncertainty modeling and hard example mining. We propose a two-stage detector named Boosting R-CNN (see Figure 2), which consists of three key components: RetinaRPN, probabilistic inference pipeline, and boosting reweighting. Specifically, RetinaRPN generates proposals from backbone features with heavier heads to perform three tasks: objectness prediction, IoU prediction, and box localization. It includes the IoU prediction and objectness as two indicators to model the prior uncertainty in order to accurately measure the vagueness of the objects. With a proposed fast IoU loss, high-quality proposals can be obtained. Second, the probabilistic inference pipeline combines the RetinaRPN's object prior and the R-CNN classification score to make a prediction, which uses the uncertainty from the first stage to improve the robustness of the detector. Third, boosting reweighting attaches more attention to hard examples whose priors are miscalculated by amplifying the loss according to the RPN's error. Since the final classification score of the object combines the RPN's prior and the R-CNN's scores, the R-CNN trained with reweighted samples has a strong robustness to hard examples, modifying its score to correct the false positive and false negative of the RetinaRPN.

With these three components, our Boosting R-CNN can handle complicated underwater challenges and be robust to vague objects. Our method is evaluated on two underwater object detection datasets: UTDAC2020[1] and Brackish [35], not only achieving state-of-the-art performance but also maintaining a relatively high inference speed. Moreover, the experiments on the Pascal VOC [13] and the MS COCO [29] dataset show Boosting R-CNN obtains favorable performance on general object detection. Our code will be released at https://github.com/mousecpn/Boosting-R-CNN-Reweighting-R-CNN-Samples-by-RPN-s-Error-for-Underwater-Object-Detection.git

## 2. Related Work

### 2.1. Object Detection

Existing object detection can be categorized into two mainstreams: two-stage and one-stage detectors. For two-stage detectors, the basic idea is to reduce the detection task to the classification problem [40]. In the first stage, the region proposal network (RPN) aims to propose candidate object bounding boxes, and RoI Pooling and RoI

Align are leveraged to crop the features from backbone and resize them to the same size. In the second stage, the R-CNN head realizes classification and regression tasks of all objects. One-stage detectors abandon the usage of the RPN and RoI Align, directly obtaining the coordinates of bounding boxes and classes of the objects. Nowadays, one-stage detectors can achieve the same level of performance as two-stage detectors. There are two branches of one-stage detectors: anchor-based methods and anchor-free methods. Early works on one-stage detectors are mostly anchor-based methods [32, 28]. Recently, some works rethink whether the anchor is necessary, and propose their designs to abandon the use of anchors [44, 59].

As the research on object detection goes deeper, the researchers find that the concepts of one-stage and two-stage detectors are not entirely different. Some research works aim to leverage the advantages of two-stage to enhance the performance of one-stage detectors. RefineDet [54] separates the one-stage detection into two sub-module: the anchor refinement module and the object detection module. AlignDet [8] uses deformable convolution (DCN) to imitate RoIAlign to obtain aligned features in the second stage. RepPoints [50] leverages the idea of refinement and feature alignment and applies it to the proposed anchor-free detectors based on keypoint detection. Two-stage detectors are also nurtured by the achievements of one-stage detectors. CenterNet2 [58] finds that a strong anchor-free one-stage detector as the RPN can predict an accurate object likelihood that informs the overall detection score. Combining the object likelihood of RPN and the conditional classification score of the R-CNN will achieve higher performance with fewer proposals, which reduces the inference cost. Our Boosting R-CNN is a probabilistic two-stage detector like CenterNet2. The difference is that we build a strong anchor-based RPN, and apply a hard example mining mechanism based on the RPN's errors.

### 2.2. Hard Example Mining.

Hard example mining methods aim to attach more attention to hard examples, relying on the hypothesis that training on hard examples leads to better performance. The first deep detector to use hard example mining is Single Shot Detector [32], which chooses only the negative examples with the highest loss values. Online Hard Exampling Mining (OHEM) [42] considers both hard positive and negative examples for training. Considering the efficiency and memory problems of OHEM, IoU-based sampling [34] is proposed, associating the hardness of the negative examples with their IoUs, and sampling averagely across all IoU ranges. Focal Loss [28] is a soft hard-example mining method, dynamically assigning more weight to the hard examples based on the classification score. Prime Sample Attention (PISA) [3] proposes an IoU Hierarchical Local Rank for all samples, assigning higher weight for positive examples with higher IoUs. Different from the methods mentioned above, our two-stage Boosting R-CNN defines the hardness of the examples based on their prior probability from the proposed

RetinaRPN. A soft reweighting mechanism is proposed to amplify the loss of the hard examples and shrink the loss of the easy examples.

### 2.3. Underwater Object Detection.

As an indispensable technology for AUVs to perform multiple tasks under the water, underwater object detection has attracted a large amount of attention from researchers all around the world. For instance, Huang et al. [19] introduce perspective transformation, turbulence simulation, and Illumination synthesis into data augmentation. Chen et al. [9] design a novel underwater salient detection model that is established by mathematically stimulating the biological vision mechanism of aquatic animals RoIMix [30] is a data augmentation method that applies mixup on the RoI level to imitate occlusion conditions. SWIPENET [7] takes full advantage of both high resolution and semantic-rich hyper feature maps to increase the performance of small objects. Besides, a novel sample-reweighted loss and a new training paradigm CMA are proposed which are noise-immune. Poisson GAN [47] is also a data augmentation method, which pastes the object on the underwater background by poisson blending and uses GAN to correct the artifact. FERNet [14] consists of three modules: composite connected backbone, receptive field augmentation module, and prediction refinement scheme. Composited FisherNet [57] is based on underwater video object detection, leveraging the differences between the foreground and background to extract salient features and proposing an enhanced path aggregation network to solve the insufficient utilization of semantic information caused by linear up-sampling. RoIAttn [27] considers RoI patches as tokens and applies the external attention module on the RoIs to improve the performance of underwater object detection. Compared with the methods mentioned above, to the best of our knowledge, our idea of considering using RPN's error for hard example mining has not been investigated by any existing underwater object detection approaches.

## 3. Boosting R-CNN

### 3.1. Overview

Different from the vanilla two-stage detector Faster R-CNN, the proposed two-stage detector Boosting R-CNN has three key components: RetinaRPN, the probabilistic inference pipeline, and boosting reweighting. The pipeline of our Boosting R-CNN is shown in Figure 2. In detail, the backbone and the feature fusion neck (e.g., ResNet+PAFPN) first extract features from images. Second, RetinaRPN provides a series of high-quality proposals with corresponding prior probability. Third, boosting reweighting amplifies the classification loss of the hard examples whose priors are miscalculated, while decreasing the weight of the easy examples with accurately estimated priors. Fourth, the R-CNN head which contains two fully-connected layers is trained on reweighted RoI samples. In the inference stage, the final score is the square root of the multiplication of the prior and the classification score.

### 3.2. Backbone and Feature Fusion Neck

Given an image $I \in \mathbb{R}^{3 \times H_0 \times W_0}$ (with RGB channels), a backbone (e.g. ResNet50) generates multi-scale feature maps $\{x^l\}_{l=3}^5$ at $C_3$-$C_5$ ($C_l$ has resolution $2^l$ smaller than the input). The multi-scale feature maps will be sent into the feature fusion neck.

PAFPN [31] is employed as the feature fusion neck. PAFPN contains two parts: the top-down path and the bottom-up path. In the top-down path, the high-level feature is used to enhance the low-level feature. Given the multi-scale feature maps $\{x^l\}_{l=3}^5$ from backbone, the output feature $\{p^l\}_{l=3}^5$ as:

$$p^5 = conv(x^5), \tag{1}$$
$$p^4 = conv(x^4) + u(p^5), \tag{2}$$
$$p^3 = conv(x^3) + u(p^4), \tag{3}$$

where $conv(\cdot)$ denotes the convolution layer, and $u(\cdot)$ denotes the 2x upsampling layer. In the bottom-up path, the low-level feature is leveraged to augment the high-level feature to obtain feature maps $\{q^l\}_{l=3}^7$, as:

$$q^3 = conv(p^3), \tag{4}$$
$$q^4 = conv(p^4) + d(q^3), \tag{5}$$
$$q^5 = conv(p^5) + d(q^4), \tag{6}$$
$$q^6 = conv_s(q^5), \tag{7}$$
$$q^7 = conv_s(q^6), \tag{8}$$

where $conv_s(\cdot)$ denotes the convolution layer with stride 2, $d(\cdot)$ denotes the 2x downsampling layer. The output multi-scale features $\{q^l\}_{l=3}^7$ are fed into the detection head.

### 3.3. RetinaRPN

The RPN is responsible for providing proposals that have potential objects. Underwater images are blurring, low-contrast and distorted, which make it difficult to distinguish the objects from the background. Besides, in the occlusion condition, the objectness trained with simple cross entropy loss in the vanilla RPN is not a good estimation of the proposal box localization accuracy. As a result, the high-quality proposals may be filtered by the poorly regressed proposals with higher objectness. To obtain high-quality proposals with accurate prior probabilities, we aim to build a strong RPN inspired by the designs of the current one-stage detector, which is named retina region proposal network (RetinaRPN).

**Heavier Head.** Instead of using one simple convolution layer in the vanilla RPN, we use four convolution layers with group normalization. More convolution layers have a more powerful capability to detect vague objects in blurring, low-contrast, and distorted underwater images.

**Multi-Ratio Anchors.** For each FPN level, we use anchors at three aspect ratios {1:2, 1:1, 2:1} with sizes {$2^0$, $2^{1/3}$, $2^{2/3}$} of $32^2$ to $512^2$ for FPN levels $Q_3$-$Q_7$. In total, there are $A$=9 anchors per pixel. Anchor is an important prior

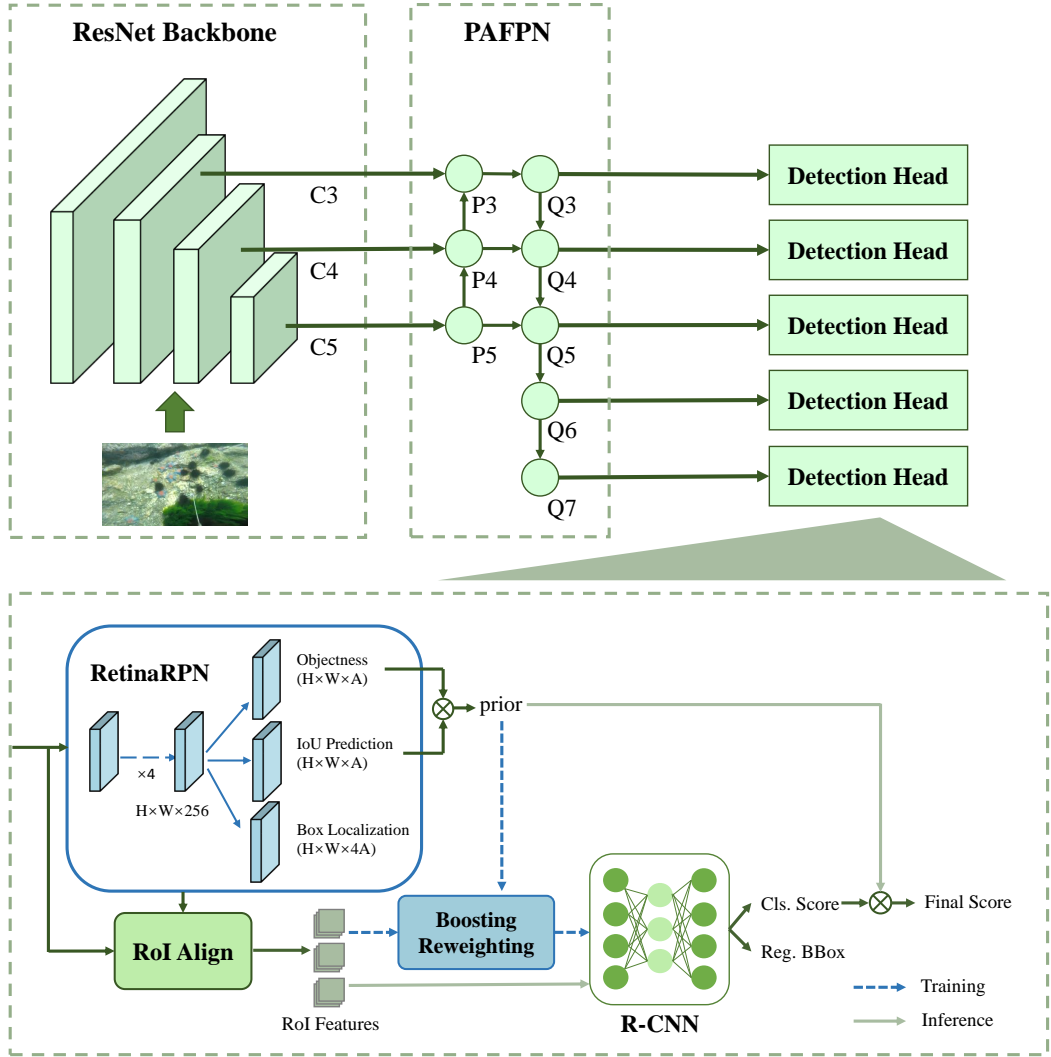

**Figure 2:** The overview of the proposed Boosting R-CNN. The backbone and the feature fusion neck first extract features from images. RetinaRPN provides a series of high-quality proposals with corresponding prior probability. Boosting reweighting amplifies the classification loss of the hard examples whose priors are miscalculated while decreasing the weight of the easy examples with accurately estimated priors. The R-CNN head which contains two fully-connected layers is trained on reweighted RoI samples. In the inference stage, the final score is the square root of the multiplication of the prior and the classification score.

for regressing and classifying aquatic organisms with vague boundaries.

**Loss Function.** RetinaRPN performs three tasks: objectness prediction, box localization, and IoU prediction. The objectness branch is trained to predict whether there is an object in an anchor. We leverage the focal loss as objectness loss:

$$L_{fl}(\hat{p}_i) = \begin{cases} -\alpha(1-\hat{p}_i)^\gamma \log(\hat{p}_i), & y_i = 1, \\ -(1-\alpha)\hat{p}_i^\gamma \log(1-\hat{p}_i), & y_i = 0, \end{cases} \quad (9)$$

$$L_{obj-rpn} = \frac{1}{n}\sum_{i=1}^{n} L_{fl}(\hat{p}_i), \quad (10)$$

where $\hat{p}_i$ is the objectness score of the anchor $i$ output by the RetinaRPN, $\alpha$ is the balance parameter, and $\gamma$ is the focus parameter. $n$ is the number of anchors. $y_i \in \{0, 1\}$ is the label of anchor $i$. It is set to 1 if anchor $i$ is a positive sample, otherwise, it is set to 0. As for the positive and negative samples assignment, the anchors with IoU over 0.5 with ground-truth boxes are regarded as positive samples, while the anchors with IoU below 0.5 are regarded as negative samples.

The localization branch aims to output the proposals which are refined on the anchors. Usually, IoU loss is leveraged in the regression loss:

$$g_i = IoU(\hat{\boldsymbol{b}_i}, \boldsymbol{b}_i^*), \quad (11)$$

$$L_{IoU}(\hat{\boldsymbol{b}_i}) = 1 - g_i, \quad (12)$$

where $\hat{\boldsymbol{b}_i}$ and $\boldsymbol{b}^*$ are the predicted box $i$ and corresponding ground-truth box, and $g_i$ is the IoU between them. IoU loss has some good properties, such as non-negative, symmetry, triangle inequality, and scale insensitivity. And it is the metric of object detection. However, the convergence speed of the IoU loss is slow. In order to increase the convergence speed, L2 loss is added to IoU loss. The improved IoU loss can be rewritten as:

$$L'_{IoU}(\hat{b}_i) = 1 - g_i + \sum_{j \in \{x,y,w,h\}} ||\hat{t_{i,j}} - t^*_{i,j}||^2_2, \quad (13)$$

$$\hat{t_{i,x}} = \frac{\hat{x}_i - x^a_i}{w^a_i}, \qquad \hat{t_{i,y}} = \frac{\hat{y}_i - y^a_i}{h^a_i}, \quad (14)$$

$$\hat{t_{i,w}} = \log(\frac{\hat{w}_i}{w^a_i}), \qquad \hat{t_{i,h}} = \log(\frac{\hat{h}_i}{h^a_i}), \quad (15)$$

$$t^*_{i,x} = \frac{x^*_i - x^a_i}{w^a_i}, \qquad t^*_{i,y} = \frac{y^*_i - y^a_i}{h^a_i}, \quad (16)$$

$$t^*_{i,w} = \log(\frac{w^*_i}{w^a_i}), \qquad t^*_{i,h} = \log(\frac{h^*_i}{h^a_i}). \quad (17)$$

where $\{x^a_i, y^a_i, w^a_i, h^a_i\}$ are the coordinates of anchor $i$, $\{\hat{x}_i, \hat{y}_i, \hat{w}_i, \hat{h}_i\}$ and $\{x^*_i, y^*_i, w^*_i, h^*_i\}$ are the coordinates of the predicted box and its corresponding ground truth, and $\{\hat{t_{i,x}}, \hat{t_{i,y}}, \hat{t_{i,w}}, \hat{t_{i,h}}\}$ and $\{t^*_{i,x}, t^*_{i,y}, t^*_{i,w}, t^*_{i,h}\}$ denote the encoding of the 4 coordinates of the predicted box and ground truth respectively. This encoding method is the same as [40]. However, L2 loss is very vulnerable to outliers, which will harm the regression accuracy. To solve this problem, we design the fast IoU loss (FIoU), which is inspired by [56], as:

$$L_{FIoU}(\hat{b}_i) = g^\eta_i (1 - g_i + \sum_{j \in \{x,y,w,h\}} ||\hat{t_{i,j}} - t^*_{i,j}||^2_2), \quad (18)$$

$$L_{loc-rpn} = \frac{1}{m} \sum_{i=1}^{m} L_{FIoU}(\hat{b}_i), \quad (19)$$

where $\eta$ is a parameter to control the degree of inhibition of outliers, and $m$ is the number of positive samples. We add an IoU weighted term $g^\eta_i$ to alleviate the problem of vulnerability to outliers. With the IoU weighted term, low-quality samples with high regression loss will be filtered, for the weighted term will become small. And the RetinaRPN will focus on the prime samples with moderate regression accuracy, which will enhance the robustness to the outliers and remain the fast convergence.

The IoU prediction branch is trained to predict the IoUs between regressed boxes and their corresponding ground truths. And the cross entropy is used as the loss function:

$$L_{iou-rpn} = \frac{1}{m} \sum_{i=1}^{m} -[g_i \log(\hat{g}_i) + (1 - g_i) \log(1 - \hat{g}_i)], \quad (20)$$

where $\hat{g}_i$ is the predicted IoU of the anchor $i$. The object prior is the square root of the multiplication of the objectness score and IoU prediction, namely:

$$pr_i = \sqrt{\hat{g}_i * \hat{p}_i}. \quad (21)$$

With the IoU prediction branch, the detector can provide the uncertainty to the prior when the objects are occluded in the underwater environment. In detail, objectness denotes the likelihood of the object in an anchor. Although objectness trained with focal loss can effectively filter the negative samples, it will also assign a high value to the proposal in which the object is severely covered by other objects. IoU prediction predicts the IoU between the proposal and its ground truth and assigns a value to the object according to its level of occlusion. Combining two indicators includes uncertainties from different perspectives, and comprehensively models the prior probabilities of the proposals.

### 3.4. Probabilistic Inference Pipeline

For the two-stage detector, in the first stage, the RPN outputs K proposal boxes $b_1, ..., b_K$. And for the proposal $k \in \{1, ..., K\}$, RPN predicts a class-agnostic foreground probability $P(O_k)$, where $O_k=1$ denotes the proposal $k$ is an object and $O_k=0$ suggests the background. This is realized by a binary classifier trained with a log-likelihood objective. In the second stage, high-scoring proposals are sampled to train the R-CNN head, a softmax classifier. The R-CNN learns to classify each proposal into one of the foreground classes or background. The output classification score of the proposal $k$ for the class $C_k$ can be seen as a conditional categorical probability $P(C_k|O_k=1)$ ($C_k \in \{C, bg\}$, $C$ is the set of classes and $bg$ denotes background). However, in the inference stage, the final detection score directly uses the classification score in the R-CNN head, ignoring the prior probability from the RPN. During the training stage in the R-CNN head, since the supervision signals of all proposals are the equivalent with a softmax classifier regardless of the localization accuracy, the R-CNN head easily outputs false over-confident predictions. Thus, compared with using the conditional categorical probability $P(C_k=c|O_k=1)$, it is more reasonable to use the marginal probability $P(C_k = c), c \in C$ as the final detection score. We set $P(C_k=bg|O_k=0)=1$ and $P(C_k = c|O_k = 0) = 0$, which means that it is impossible for the R-CNN head to reconsider a proposal as a positive sample if the RPN regards the proposal as a negative sample. The marginal probability $P(C_k = c)$ can be written as:

$$\begin{aligned} P(C_k = c) &= \sum_{u \in \{0,1\}} P(C_k = c|O_k)P(O_k = u) \\ &= P(C_k = c|O_k = 1)P(O_k = 1) \\ &\quad + P(C_k = c|O_k = 0)P(O_k = 0) \\ &= P(C_k = c|O_k = 1)P(O_k = 1). \end{aligned} \quad (22)$$

From the equation above, the marginal probability is the multiplication of the first-stage prior probability and the second-stage conditional distribution. Using the marginal probability, the first-stage prior uncertainty can be taken into consideration in the final prediction. In the implementation, the final detection score of the proposal $k$ is the square root of the multiplication of RetinaRPN's prior and the R-CNN's

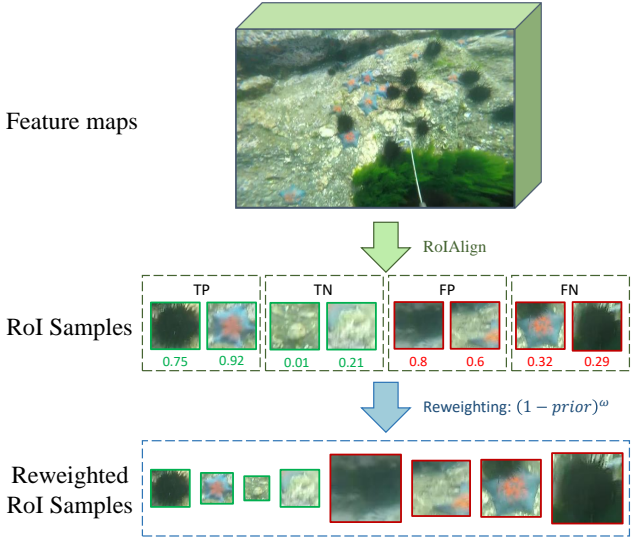

Feature maps

RoIAlign

RoI Samples

TP TN FP FN

0.75 0.92 0.01 0.21 0.8 0.6 0.32 0.29

Reweighting: $(1 - prior)^\omega$

Reweighted RoI Samples

**Figure 3:** The overview of the proposed boosting reweighting. The patch size denotes the weight of RoI samples.

classification score, namely:

$$s_k(c) = \sqrt{pr_k * cls_k(c)}, \tag{23}$$

where $cls_k(c)$ is the classification score of the sample $k$ for class $c$ in R-CNN, and $s_k(c)$ is the final score.

With the probabilistic inference pipeline, the detector can take the first-stage uncertainty into consideration to make the final predictions. Thus, compared with using the conditional probability, the marginal probability is a better estimation of the detection box localization accuracy.

### 3.5. Boosting Reweighting

There is a deficiency in the previous probabilistic inference pipeline. In the original two-stage detector, the second stage makes predictions that is independent of the first stage. As a result, a low score for a high-quality sample in the first stage will not influence the final detection result as long as the sample is selected as a proposal. However, in the probabilistic two-stage pipeline, when the RPN mistakenly generates a low prior for a high-quality positive proposal, it is hard to re-consider it as a high-confidence prediction, for the final score is the square root of the multiplication of prior and classification score. In underwater environments, vague objects as hard examples often happen, and the RPN will severely suffer from those.

To solve this problem, we hope that when the RPN miscalculates the prior of the proposal, the R-CNN can rectify the error. Thus, we propose a soft sampling strategy named boosting reweighting (BR, shown in Fig. 3), which borrows the idea of reweighting from boosting algorithm and well fits the existing frameworks. Different from vanilla Faster R-CNN, where the weights of all proposals are set to 1, BR tends to attach more attention to hard examples whose priors are miscalculated. In detail, for the sample $k$,

its classification weight is:

$$w_k = \begin{cases} (1 - pr_k)^\omega, & k \in \mathcal{F}, & (24a) \\ pr_k^\omega, & k \in \mathcal{B}, & (24b) \end{cases}$$

where $\omega \geq 0$ is the boosting parameter. $\mathcal{F}$ denotes the set of foreground samples, and $\mathcal{B}$ denotes the set of background samples. With the reweighting scheme, the classification loss of the R-CNN can be written as:

$$L_{cls} = \frac{1}{K} \sum_{k=1}^{K} w_k \cdot \sum_{c=1}^{C} (-s_k^c \cdot \log(\hat{s_k^c})), \tag{25}$$

where $\hat{s_k^c}$ and $s_k^c$ denote the predicted classification score and label of sample $k$ for class $c$, $s_k^c \in \{0, 1\}$. $K$ and $C$ are the number of proposals in second stage and the number of classes respectively. Note that the weighted terms are all smaller than 1, the total value of the classification loss will shrink, which will cause the shrink of the gradient. In order to keep the norm of the total loss unchanged, we normalize $w$ to $w'$:

$$w_k' = w_k \cdot \frac{\sum_{k=1}^{K} \sum_{c=1}^{C} (-s_k^c \cdot \log(\hat{s_k^c}))}{\sum_{k=1}^{K} w_k \cdot \sum_{c=1}^{C} (-s_k^c \cdot \log(\hat{s_k^c}))}, \tag{26}$$

$$L_{cls}' = \frac{1}{K} \sum_{k=1}^{K} w_k' \cdot \sum_{c=1}^{C} (-s_k^c \cdot \log(\hat{s_k^c})). \tag{27}$$

When the detector encounters hard positive/negative samples, obviously the priors of RPN will be small/large. As a result, the weighted term $(1 - prior(k))^\gamma / prior(k)^\gamma$ will increase and amplify the loss of the hard examples, while the loss of the easy samples will be decreased.

BR can be seen as hard example mining. There are two similar works to our BR: OHEM and focal loss. OHEM is a bootstrapping method, which is originally designed for Fast R-CNN (without RPN), it performs a feedforward for all RoIs on the R-CNN, and selects the hardest samples for training on the second feedforward. Our BR leverages the prior information from the RPN from only one feedforward, saving lots of memory cost and training time. Focal loss is designed for RetinaNet to solve the extreme imbalance between foreground and background. However, the NMS in the RPN and bootstrapping mechanism in the second stage alleviate the imbalance problem, which overlaps the function of focal loss. Our BR is used in combination with NMS and bootstrapping. To avoid the shrinking of the loss, normalization is leveraged to re-distribute the weight of each sample. Thus, BR aims to handle the problem of the hard samples in the underwater environment instead of foreground-background imbalance. Both OHEM and focal loss reweight the loss by the R-CNN's error, while our BR reweights the loss by the RPN's error. The R-CNN trained with BR excavates the subtle differences between aquatic organisms and background and is robust to the samples to which the RPN is vulnerable. Thus, the R-CNN can rectify the RPN's error in the inference stage. The experiments show that BR is more compatible with the probabilistic inference pipeline compared with other methods.

**Table 1**
Comparisons with other object detection methods on UTDAC2020 dataset. The FPS is tested on a single Nvidia GTX 1080Ti GPU. '*' means that the model uses the second training recipe.

| Method | Backbone | AP | AP50 | AP75 | APS | APM | APL | FPS |
|---|---|---|---|---|---|---|---|---|
| **Two-Stage Detector:** | | | | | | | | |
| Faster R-CNN w/ FPN [40] | ResNet50 | 44.5 | 80.9 | 44.1 | 20.0 | 39.0 | 50.8 | 11.6 |
| OHEM+Faster R-CNN w/ FPN [42] | ResNet50 | 45.1 | 82.0 | 45.1 | 21.6 | 39.1 | 51.4 | 11.6 |
| Cascade R-CNN [2] | ResNet50 | 46.6 | 81.5 | 49.3 | 21.0 | 40.9 | 53.3 | 8.8 |
| Libra R-CNN [34] | ResNet50 | 45.8 | 82.0 | 46.4 | 20.1 | 40.2 | 52.3 | 11.0 |
| Cascade RPN [45] | ResNet50 | 46.5 | 79.5 | 41.2 | 20.4 | 38.6 | 47.7 | 8.3 |
| Faster R-CNN w/ PAFPN [31] | ResNet50 | 45.5 | 82.1 | 45.9 | 18.8 | 39.7 | 51.9 | 10.9 |
| Double-Head [48] | ResNet50 | 45.3 | 81.5 | 45.7 | 20.2 | 40.0 | 51.4 | 5.7 |
| Dynamic R-CNN [51] | ResNet50 | 45.6 | 80.1 | 47.3 | 19.0 | 39.7 | 52.1 | 12.1 |
| Faster R-CNN w/ FPG [5] | ResNet50 | 45.4 | 81.6 | 46.0 | 19.8 | 39.7 | 51.4 | 13.1 |
| GRoIE [41] | ResNet50 | 45.7 | 82.4 | 45.6 | 19.9 | 40.1 | 52.0 | 6.0 |
| SABL+Faster R-CNN [46] | ResNet50 | 46.6 | 81.6 | 48.2 | 19.6 | 40.4 | 53.4 | 9.9 |
| PISA [3] | ResNet50 | 46.3 | 82.1 | 47.4 | 20.8 | 40.8 | 52.6 | 10.3 |
| Sparse R-CNN [43] | ResNet50 | 37.4 | 70.4 | 35.9 | 17.7 | 33.3 | 43 | 10.8 |
| DetectoRS [36] | ResNet50 | 47.6 | 82.8 | 49.9 | 23.1 | 41.8 | 54.2 | 4.0 |
| RoIAttn [27] | ResNet50 | 46.0 | 82.0 | 47.5 | 22.9 | 40.5 | 52.2 | 8.8 |
| CenterNet2 [58] | ResNet50 | 47.2 | 81.6 | 49.8 | 18.2 | 41.3 | 53.4 | 14.2 |
| CenterNet2* [58] | ResNet50 | 48.9 | 83.0 | 52.6 | 21.7 | 43.5 | 55.2 | 14.2 |
| **One-Stage Detector:** | | | | | | | | |
| SSD512 [32] | VGG16 | 40.0 | 77.5 | 36.5 | 14.7 | 36.1 | 45.1 | **25.0** |
| RetinaNet [28] | ResNet50 | 43.9 | 80.4 | 42.9 | 18.1 | 38.2 | 50.1 | 11.4 |
| FSAF [61] | ResNet50 | 43.9 | 81.0 | 42.9 | 18.5 | 38.9 | 50.9 | 12.8 |
| CenterNet [59] | ResNet18 | 31.3 | 61.1 | 27.6 | 11.9 | 32.5 | 33.4 | 6.2 |
| FCOS [44] | ResNet50 | 43.9 | 81.1 | 43.0 | 19.9 | 38.2 | 50.4 | 12.7 |
| RepPoints [50] | ResNet50 | 44.0 | 80.5 | 43.0 | 18.7 | 38.5 | 50.3 | 11.1 |
| FreeAnchor [55] | ResNet50 | 46.3 | 82.3 | 46.9 | 21.0 | 40.5 | 52.6 | 11.4 |
| RetinaNet w/ NASFPN [17] | ResNet50 | 37.4 | 70.3 | 35.8 | 12.4 | 36.4 | 40.4 | 13.8 |
| ATSS [53] | ResNet50 | 46.2 | 82.5 | 46.9 | 19.7 | 41.4 | 52.4 | 11.8 |
| PAA [20] | ResNet50 | 47.5 | 83.1 | 49.7 | 19.5 | 42.4 | 53.6 | 6.6 |
| AutoAssign [60] | ResNet50 | 46.3 | 83.0 | 47.6 | 18.0 | 41.3 | 52.2 | 12.3 |
| GFL [25] | ResNet50 | 46.4 | 81.9 | 47.8 | 19.3 | 40.9 | 52.5 | 12.7 |
| VFNet [52] | ResNet50 | 44.0 | 79.3 | 44.1 | 18.8 | 38.1 | 50.4 | 10.5 |
| **Transfromer:** | | | | | | | | |
| Deformable DETR [62] | ResNet50 | 46.6 | 84.1 | 47.0 | 24.1 | 42.4 | 51.9 | 7.6 |
| **Ours:** | | | | | | | | |
| **Boosting R-CNN** | ResNet50 | 48.5 | 82.4 | 52.5 | 21.1 | 42.4 | 55.0 | 13.5 |
| **Boosting R-CNN*** | ResNet50 | **51.4** | **85.5** | **56.8** | **23.8** | **45.8** | **57.8** | 13.5 |

## 3.6. Loss Function

In the R-CNN head, we apply L1 loss on the positive proposals in the second stage for regression:

$$L_{reg} = \frac{1}{K_{pos}} \sum_{k=1}^{K_{pos}} \sum_{j \in \{x,y,w,h\}} |\hat{t_{k,j}} - t^*_{k,j}|, \quad (28)$$

$$\hat{t_{k,x}} = \frac{\hat{x_k} - x^p_k}{w^p_k}, \ \hat{t_{k,y}} = \frac{\hat{y_k} - y^p_k}{h^p_k}, \quad (29)$$

$$\hat{t_{k,w}} = \log(\frac{\hat{w_k}}{w^p_k}), \ \hat{t_{k,h}} = \log(\frac{\hat{h_k}}{h^p_k}), \quad (30)$$

$$t^*_{k,x} = \frac{x^*_k - x^p_k}{w^p_k}, \ t^*_{k,y} = \frac{y^*_k - y^p_k}{h^p_k}, \quad (31)$$

$$t^*_{k,w} = \log(\frac{w^*_k}{w^p_k}), \ t^*_{k,h} = \log(\frac{h^*_k}{h^p_k}), \quad (32)$$

where $\{x^p_k, y^p_k, w^p_k, h^p_k\}$ are the coordinates of proposals $k$, $\{\hat{x_k}, \hat{y_k}, \hat{w_k}, \hat{h_k}\}$ and $\{x^*_k, y^*_k, w^*_k, h^*_k\}$ are the coordinates of predicted box and its corresponding ground truths, $\{\hat{t_{k,x}}, \hat{t_{k,y}}, \hat{t_{k,w}}, \hat{t_{k,h}}\}$ and $\{t^*_{k,x}, t^*_{k,y}, t^*_{k,w}, t^*_{k,h}\}$ denote the encoding of the 4 coordinates of the predicted box and ground truth respectively. The total loss in the Boosting R-CNN is:

$$L_{total} = \lambda_{obj-rpn}L_{obj-rpn} + \lambda_{loc-rpn}L_{loc-rpn} \\ + \lambda_{iou-rpn}L_{iou-rpn} + \lambda_{reg}L_{reg} + \lambda_{cls}L_{cls}, \quad (33)$$

where $\lambda_{obj-rpn}$, $\lambda_{loc-rpn}$, $\lambda_{iou-rpn}$, $\lambda_{reg}$, $\lambda_{cls}$ are the balanced parameters for each loss respectively.

**Table 2**
Comparisons with other object detection methods on Brackish dataset. "Baseline" is the performance reported in the original paper of Brackish.

| Method | Backbone | AP | AP50 |
|---|---|---|---|
| Baseline (YOLOv3) [35] | DarkNet53 | 38.9 | 83.7 |
| Faster R-CNN w/ FPN [21] | ResNet50 | 79.3 | **97.4** |
| Cascade R-CNN [2] | ResNet50 | 80.7 | 96.9 |
| RetinaNet [28] | ResNet50 | 78.0 | 96.5 |
| DetectoRS [36] | ResNet50 | 81.6 | 97.0 |
| CenterNet2 [58] | ResNet50 | 79.3 | **97.4** |
| **Boosting R-CNN** | ResNet50 | **82.0** | **97.4** |

# 4. Experiments

## 4.1. Datasets

We conduct experiments on four challenging object detection datasets to validate the generalization performance of our method.

**(i) UTDAC2020** is an underwater dataset from the underwater target detection algorithm competition 2020. There are 5,168 training images and 1,293 validation images. It contains four classes: echinus, holothurian, starfish, and scallop. The images contain four resolutions: 3840×2160, 1920×1080, 720×405, and 586×480. We follow the COCO-style evaluation metric.

**(ii) Brackish** is an early proposed underwater image dataset collected in temperate brackish waters. It contains six classes: bigfish, crab, jellyfish, shrimp, small fish, and starfish. There are 9,967, 1,467, 1,468 images in training, validation, and test set, containing 25,613 annotations. The image size is 960×540. We follow the MS COCO-style AP[0.5:0.95:0.05] metric and Pascal VOC-style AP50 metric as the original paper.

**(iii) Pascal VOC** is a generic object detection dataset, which contains 20 object categories. The dataset includes VOC2007 part and VOC2012 part. In VOC2007 part, there are 9963 annotated images, consisting of trainval (5011 images) and test set (4952 images). In VOC2012 set, there are 11540 annotated images in trainval set. We train our detector on 07+12 trainval dataset, and evaluate it on 07 test set.

**(iiii) MS COCO** is a generic object detection dataset, which contains 80 object categories. It contains 118k images for training (trainval), 5k images for validation (val) and 20k images for testing without provided annotations (test-dev). The final results are reported on test-dev set.

## 4.2. Implementation Details

Our method is implemented on MMdetection [6]. There are two training recipes in the experiments. The first one is the default training recipes, which adopts the classic 1x training scheme (12 epochs). SGD is adopted as an optimizer, where the weight decay is 0.0001 and the momentum is 0.9. The initial learning rate is 0.005. The learning rate is divided by a factor of 10 at epoch 8 and 11. No extra data augmentation except the traditional horizontal flipping is utilized. The second training recipe adopts the 3x training scheme (36 epochs) with crop and multi-scale augmentation. AdamW is leveraged as the optimizer with an initial learning rate of 0.0001 and a weight decay of 0.05. The learning rate is divided by a factor of 10 at epoch 24 and 33.

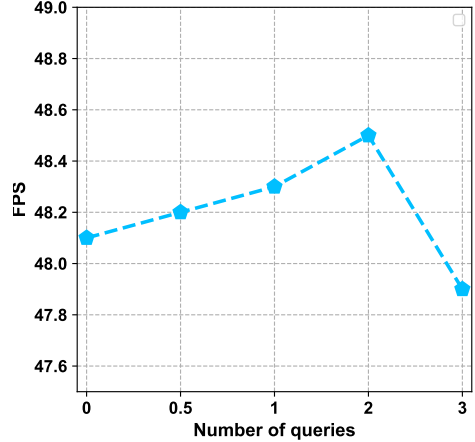

**Figure 4:** The choice of $\eta$ in FIoU loss. $\eta = 0$ means droping the IoU weighted term.

The method is trained on a single NVIDIA GTX 1080Ti GPU. During the inference, we use a maximum of 256 proposal boxes in the second stage, which improves the inference speed. As for the balanced parameters of loss, $\lambda_{obj-rpn}$, $\lambda_{loc-rpn}$, $\lambda_{iou-rpn}$, $\lambda_{cls}$, $\lambda_{reg}$ are set to 1, 2, 1, 2, 2 respectively.

## 4.3. Comparisons with Other State-of-the-art Methods

We compare Boosting R-CNN against some state-of-the-art methods in four object detection datasets in Table 1, 2, 3, and 4.

### 4.3.1. Results on UTDAC2020

The experiment results on the UTDAC2020 dataset are shown in Table 1. CenterNet2* and Boosting R-CNN* denote that the models use multi-scale training and 3× training time. Besides, Deformable DETR is trained for 50 epochs with multi-scale training. All the detectors are implemented by MMdetection [6] except CenterNet2 which is officially implemented in Detectron2 [49].

As shown in Table 1, in single-scale training setting, Boosting R-CNN achieves 48.5% AP, which is higher than DetectoRS (47.6% AP), PAA (47.5% AP) and CenterNet2 (47.2% AP). In multi-scale training setting, Boosting R-CNN still surpasses CenterNet2 (51.4% AP vs 48.9% AP). As a result, our Boosting R-CNN defeats all the detectors and builds new state-of-the-art performance. As for the inference speed, Boosting R-CNN achieves 13.5 FPS, which is higher than most of the two-stage detectors including Faster R-CNN (11.6 FPS) but lower than CenterNet2 (14.2 FPS).

**Table 3**
Comparisons with other object detection methods on PASCAL VOC dataset. '*' means that our model uses the second training recipe. The performances of Faster R-CNN and RetinaNet are published by MMdetection. "†" means our re-implementation. The performances of other methods are from their original paper.

| Method | Backbone | Input Size | mAP |
|---|---|---|---|
| **Two-Stage Detector:** | | | |
| Faster R-CNN w/ FPN [21] | ResNet50 | $1000 \times 600$ | 79.5 |
| MR-CNN [18] | VGG16 | $1000 \times 600$ | 78.2 |
| R-FCN [10] | ResNet101 | $1000 \times 600$ | 80.5 |
| RON384++ [22] | VGG16 | $384 \times 384$ | 77.6 |
| Cascade R-CNN† [2] | ResNet50 | $1000 \times 600$ | 80.0 |
| CenterNet2† [58] | ResNet50 | $1000 \times 600$ | 76.8 |
| **One-Stage Detector:** | | | |
| SSD300 [32] | VGG16 | $300 \times 300$ | 74.3 |
| SSD512 [32] | VGG16 | $512 \times 512$ | 76.8 |
| YOLO [38] | GoogleNet | $448 \times 448$ | 63.4 |
| YOLOv2 [39] | DarkNet19 | $544 \times 544$ | 78.6 |
| RefineDet512 [54] | VGG16 | $512 \times 512$ | 81.8 |
| DSSD513 [15] | ResNet101 | $513 \times 513$ | 81.5 |
| RetinaNet [28] | ResNet50 | $1000 \times 600$ | 77.3 |
| FERNet [14] | VGG16+ResNet50 | $512 \times 512$ | 81.0 |
| **Boosting R-CNN** | ResNet50 | $1000 \times 600$ | 81.9 |
| **Boosting R-CNN*** | ResNet50 | $1000 \times 600$ | **83.0** |

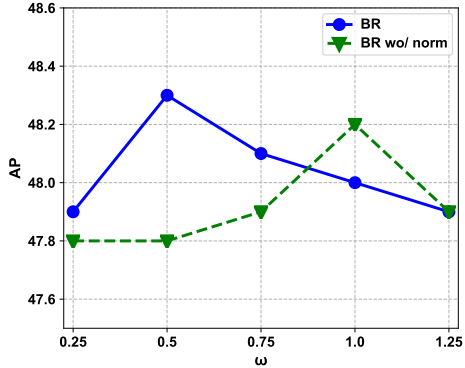

**Figure 5:** The choice of $\omega$ in boosting reweighting. Blue line denotes the BR with normalization, while green dash line denotes the BR without normalization.

Moreover, among the one-stage detectors, we can conclude that the anchor-based methods (SSD, RetinaNet, ATSS, FreeAnchor, PAA, and GFL) can obtain relatively high performances than the anchor-free methods (FSAF, FCOS, RepPoints, VFNet, and AutoAssign). That is because the boundaries of the aquatic organisms are vague in the low-contrast and distorted underwater images, and anchors help the model to obtain boundaries prior, which boosts the convergence.

### 4.3.2. Results on Brackish

The experiment results on the Brackish dataset are shown in Table 2. Boosting R-CNN achieves 82.0% AP and 97.4% AP50, achieving the highest performance. Compared with DetectoRS (81.6% AP and 97.4% AP50), Boosting R-CNN is 0.4% higher in both AP and AP50 metric.

### 4.3.3. Results on Pascal VOC

The experiment results on Pascal VOC are shown in Table 3. In single-scale training setting, Boosting R-CNN achieves 81.9% mAP performance, which is higher than Faster R-CNN (79.5%), Cascade R-CNN (80.0%), DSSD513 (81.5% AP) and CenterNet2 (76.8%). In multi-scale training setting, Boosting R-CNN achieves the highest performance of 83.0% mAP.

### 4.3.4. Results on MS COCO

Table 4 compares our method to the state-of-the-art detectors with a large backbone on MS COCO test-dev. All the models presented are using single-scale testing. Boosting R-CNN achieves 44.4% AP only with ResNet50 backbone, outperforming some two-stage detectors (e.g. Cascade R-CNN, Grid R-CNN, and Libra R-CNN), and some one-stage detectors (e.g. FCOS, CornerNet, and FSAF). And Boosting R-CNN further achieves 50.7% AP with Res2Net101-DCN backbone, outperforming GFLV2 with the same backbone (50.6% AP) and achieving the highest performance.

## 4.4. Ablation Study

We also conduct extensive ablation studies on UTDAC2020 to validate each module in our proposed Boosting R-CNN. Table 5 shows the detailed road map from the default Faster R-CNN to the proposed Boosting R-CNN. Our proposed RetinaRPN improves the performance from 44.5% AP (row 1) to 46.9% AP (row 13), which indicates that RetinaRPN provides higher-quality proposals. With an accurate estimation of the object prior in the RetinaRPN, the probabilistic inference pipeline considers the uncertainty of the first stage to make a prediction, increasing the performance from 46.9% AP to 47.9% AP (row 13 vs row 14). Using the boosting reweighting alone achieves a favorable improvement (from 44.5% AP to 45.3% AP), which is higher than OHEM (45.1% AP) in Table 1, and BR boosts the performance from 47.9% AP to 48.3% AP (row 14 vs row 15). With PAFPN, the final performance of 48.5% AP can be obtained (row 16).

### 4.4.1. Ablation Study of RetinaRPN

The first 5 columns in Table 5 shows the detailed ablation studies of the proposed RetinaRPN. (i) Using 4 layer convolution layers with GN (row 1 vs row 2) increases the capability of feature extraction, improving performance (44.5% to 45.1% in AP). (ii) Using focal loss can take all the samples into consideration (row 4), and further improve the performance (45.4 % AP). (iii) Compared with 3 anchors, 9 anchors provide a great performance increase (row 4 vs row 6, 45.4% to 46.7% AP), which further illustrates the importance of anchors in underwater environments. (vi) Adding an IoU prediction provides an accurate prior, which further improves the robustness to the vague objects (row 8 vs row 9, 47.2% AP to 47.5% AP). (vi) Our proposed fast IoU loss (row 14, 47.9% AP) is superior to L1 loss (row 9, 47.5% AP), GIoU loss (row 10, 47.6% AP), CIoU loss (row 11, 47.6% AP), and focal EIoU loss (row12, 47.7% AP).

**Table 4**
Comparisons with other object detection methods on MS COCO dataset with the large backbone in single-scale testing. '*' means that our model uses the second training recipe. The performances of other methods are published in their papers, and they all use 3x training time.

| Method | Backbone | AP | AP50 | AP75 | APS | APM | APL |
|--------|----------|-----|------|------|-----|-----|-----|
| **Two-Stage Detector:** | | | | | | | |
| Faster R-CNN w/ FPN[40] | ResNet101 | 36.2 | 59.1 | 39.0 | 18.2 | 39.0 | 48.2 |
| Cascade R-CNN [2] | ResNet101 | 42.8 | 62.1 | 46.3 | 23.7 | 45.5 | 55.2 |
| Grid R-CNN [33] | ResNet101 | 41.5 | 60.9 | 44.5 | 23.3 | 44.9 | 53.1 |
| Libra R-CNN [34] | ResNeXt101-64x4d | 43.0 | 64.0 | 47.0 | 25.3 | 45.6 | 54.6 |
| Double-Head [48] | ResNet101 | 42.3 | 62.8 | 46.3 | 23.9 | 44.9 | 54.3 |
| Dynamic R-CNN [51] | ResNet101-DCN | 46.9 | 65.9 | 51.3 | 28.1 | 49.6 | 60.0 |
| BorderDet [37] | ResNeXt101-64x4d-DCN | 48.0 | 67.1 | 52.1 | 29.4 | 50.7 | 60.5 |
| TridentNet [26] | ResNet101-DCN | 46.8 | 67.6 | 51.5 | 28.0 | 51.2 | 60.5 |
| CPN [12] | HG104 | 47.0 | 65.0 | 51.0 | 26.5 | 50.2 | 60.7 |
| **One-Stage Detector:** | | | | | | | |
| FCOS [44] | ResNeXt101-64x4d-DCN | 46.6 | 65.9 | 50.8 | 28.6 | 49.1 | 58.6 |
| CornerNet [23] | HG104 | 40.6 | 56.4 | 43.2 | 19.1 | 42.8 | 54.3 |
| CenterNet [59] | HG104 | 42.1 | 61.1 | 45.9 | 24.1 | 45.5 | 52.8 |
| CentripetalNet [11] | HG104 | 46.1 | 63.1 | 49.7 | 25.3 | 48.7 | 59.2 |
| RetinaNet [28] | ResNet101 | 39.1 | 59.1 | 42.3 | 21.8 | 42.7 | 50.2 |
| FSAF [61] | ResNeXt101-64x4d | 42.9 | 63.8 | 46.3 | 26.6 | 46.2 | 52.7 |
| RepPoints [50] | ResNet101-DCN | 45.0 | 66.1 | 49.0 | 26.6 | 48.6 | 57.5 |
| RepPointsV2 | ResNet101-DCN | 48.1 | 67.5 | 51.8 | 28.7 | 50.9 | 60.8 |
| FreeAnchor [55] | ResNeXt101-32x8d | 46.0 | 65.6 | 49.8 | 27.8 | 49.5 | 57.7 |
| ATSS [53] | ResNeXt101-32x8d-DCN | 47.7 | 66.5 | 51.9 | 29.7 | 50.8 | 59.4 |
| PAA [20] | ResNeXt101-64x4d-DCN | 49.0 | 67.8 | 53.3 | 30.2 | 52.8 | 62.2 |
| AutoAssign [60] | ResNeXt101-64x4d-DCN | 49.5 | 68.7 | 54.0 | 29.9 | 52.6 | 62.0 |
| GFL [25] | ResNeXt101-32x4d-DCN | 48.2 | 67.4 | 52.6 | 29.2 | 51.7 | 60.2 |
| GFLV2 [24] | Res2Net101-DCN | 50.6 | 69.0 | 55.3 | 31.3 | 54.3 | 63.5 |
| YOLOv4 [1] | CSPDarkNet-53 | 43.5 | 65.7 | 47.3 | 26.7 | 46.7 | 53.3 |
| **Transformer:** | | | | | | | |
| DETR [4] | ResNet101 | 43.5 | 63.8 | 46.4 | 21.9 | 48.0 | 61.8 |
| Deformable DETR [62] | ResNeXt101-64x4d-DCN | 50.1 | **69.7** | 54.6 | 30.6 | 52.8 | **65.6** |
| **Ours:** | | | | | | | |
| **Boosting R-CNN\*** | ResNet50 | 44.4 | 63.9 | 48.2 | 26.9 | 47.0 | 54.8 |
| **Boosting R-CNN\*** | Res2Net101-DCN | **50.7** | 69.2 | **55.8** | **31.7** | **54.1** | 63.5 |

Figure 4 shows the choice of hyper-parameter $\eta$ in fast IoU loss. If $\eta$ is too large, the gradient will be dominated by the easy samples with high IoUs. If $\eta$ is too small, it will lack the ability to filter the outliers. When $\eta$ is set to 2, the highest performance of 48.5% AP can be obtained. When $\eta$ is set to 0, which is equivalent to the fast IoU loss without the IoU weighted term, the performance is lower. Thus, the training of the model will suffer from the outliers.

### 4.4.2. Hard Example Mining

Table 6 shows the experiments of different hard example mining methods. Since our BR is a kind of hard example mining, it is necessary to compare it to other hard example mining methods, i.e., OHEM, PISA, and focal loss. We replace BR with these methods to evaluate the effectiveness and the compatibility with the probabilistic inference pipeline. "Cls. Loss" denotes the classification loss in the R-CNN head, "Random" means randomly sampling positive and negative RoIs during training. The first row (47.9% AP) corresponds to the next-to-last row in Table 5. BR achieves the highest performance (48.3% AP). Using OHEM instead gives a relatively lower performance (47.5% AP). PISA severely does harm to the performance (46.9% AP). Focal loss also causes severe performance decrease. When the $\gamma$ is set lower, which means that focal loss gets closer to cross-entropy loss, the performance is restored. The experiments in 6 show that in the probabilistic pipeline, BR helps R-CNN to correct the mistakes of RPN, which is more compatible than OHEM, PISA, and focal loss.

Figure 5 is the experiment of the choice of $\omega$ in BR. In this experiment, PAFPN is not used. From the figure, it can be concluded that normalization improves the performance and shifts the optimal value of $\omega$. The highest performance of 48.3% AP is obtained when $\omega$ is set to 0.5.

### 4.4.3. Anchor Assignment

In Table 7, we adopt other positive and negative anchor assignment strategies in our RetinaRPN. PAFPN and boosting reweighting is not leveraged in the experiments. Although ATSS [53], PAA [20] and OTA [16] achieve astonishing performances in one-stage detectors, they decreases the performance when they play the role of the RPN. The reason may be that the adaptive assignment provides over-confident priors, which decreases the recall. Our setting "(0.5, 0.5)" achieves the best performance in underwater object detection.

**Table 5**
A detailed abalation study of Boosting R-CNN. The first five columns denotes the ablation studies of RetinaRPN. "4 L." denotes using 4 convolution layers with GN. "NA." denotes the number of anchors. "Reg Loss" denotes the regression loss in RPN. "FL" denotes using focal loss and abandoning the bootstrapping in RPN. "IoUp." means adding IoU prediction in RPN with cross entropy loss. "Prob" denotes using probabilistic inference pipeline. "BR" means boosting reweighting.

| Row | RetinaRPN | | | | | Neck | Prob | BR | AP | AP50 | AP75 |
| | 4 L. | NA. | Reg Loss | FL | IoUp. | | | | | | |
|---|---|---|---|---|---|---|---|---|---|---|---|
| 1 | | 3 | L1 | | | FPN | | | 44.5 | 80.9 | 44.1 |
| 2 | ✓ | 3 | L1 | | | FPN | | | 45.1 | 81.6 | 45.9 |
| 3 | | 3 | L1 | | | FPN | | ✓ | 45.3 | 81.6 | 45.8 |
| 4 | ✓ | 3 | L1 | ✓ | | FPN | | | 45.4 | 81.2 | 46.3 |
| 5 | ✓ | 3 | GIoU | ✓ | | FPN | | | 45.8 | 80.5 | 47.5 |
| 6 | ✓ | 9 | L1 | ✓ | | FPN | | | 46.7 | 80.0 | 49.0 |
| 7 | ✓ | 9 | GIoU | ✓ | | FPN | ✓ | | 46.8 | 82.2 | 48.8 |
| 8 | ✓ | 9 | L1 | ✓ | | FPN | ✓ | | 47.2 | 82.5 | 49.3 |
| 9 | ✓ | 9 | L1 | ✓ | ✓ | FPN | ✓ | | 47.5 | **83.0** | 50.3 |
| 10 | ✓ | 9 | GIoU | ✓ | ✓ | FPN | ✓ | | 47.6 | 82.7 | 50.4 |
| 11 | ✓ | 9 | CIoU | ✓ | ✓ | FPN | ✓ | | 47.6 | 82.8 | 50.1 |
| 12 | ✓ | 9 | F-EIoU | ✓ | ✓ | FPN | ✓ | | 47.7 | **83.0** | 49.8 |
| 13 | ✓ | 9 | FIoU | ✓ | ✓ | FPN | | | 46.9 | 81.3 | 49.6 |
| 14 | ✓ | 9 | FIoU | ✓ | ✓ | FPN | ✓ | | 47.9 | 82.8 | 50.7 |
| 15 | ✓ | 9 | FIoU | ✓ | ✓ | FPN | ✓ | ✓ | 48.3 | 82.6 | 51.8 |
| 16 | ✓ | 9 | FIoU | ✓ | ✓ | PAFPN | ✓ | ✓ | **48.5** | 82.4 | **52.5** |

**Table 6**
The ablation studies of hard example mining. "FL$(\alpha,\gamma)$" denotes using focal loss with hyper-parameter $\alpha$ and $\gamma$.

| Cls. Loss | Sampling | AP | AP50 | AP75 |
|---|---|---|---|---|
| CE | Random | 47.9 | 82.8 | 50.7 |
| CE | OHEM | 47.5 | 82.1 | 49.8 |
| CE | PISA | 46.9 | 82.7 | 48.2 |
| FL (0.25, 2) | Random | 44.7 | 79.7 | 45.8 |
| FL (0.5, 1) | Random | 46.5 | 81.1 | 48.8 |
| FL (0.5, 0.1) | Random | 47.0 | 82.0 | 48.9 |
| FL (0.25, 2) | None | 46.7 | 80.2 | 49.7 |
| CE | **BR** | **48.3** | **82.6** | **51.8** |

**Table 7**
Positive and negative assignment. "(0.5, 0.5)" denotes that the samples with IoUs over 0.5 are regarded as positive, while the samples with IoUs below 0.5 are regarded as negative. BR is not used in the experiment.

| Assignment | AP | AP50 | AP75 |
|---|---|---|---|
| ATSS | 46.5 | 81.9 | 48.1 |
| PAA | 47.3 | 83.1 | 49.1 |
| OTA | 46.4 | 81.1 | 48.8 |
| (0.5, 0.5) | **47.9** | **82.8** | **50.7** |

### 4.5. Qualitative Comparisons

Figure 6 shows the qualitative comparison between Boosting R-CNN and other state-of-the-art methods on UTDAC2020 dataset. We apply the detectors to some challenging cases, and the prediction score threshold set to 0.05. For clarity, in each image, we visualize the prediction boxes with the top-k scores, and k is the number of the ground-truth boxes in the images. The orange boxes in the images denote a prediction box whose IoU with a certain ground truth is over 0.5 and higher than other predictions. The blue boxes denote the unmatched predictions. Besides, we also print the missed ground-truth boxes in red. Thus, more blue boxes in the images suggest lower precision, and more red boxes in the images suggest lower recall.

The first two rows denote the blurring and low contrast conditions. Boosting R-CNN can detect all the ground truths (no red box) in the images with the highest precision (only one blue box). The third row denotes the unbalanced light condition. ATSS, PAA, and DetectoRS all miss the echinus in the center. Our Boosting R-CNN does not miss any ground truths. The Fourth row denotes the occlusion condition. Boosting R-CNN can detect the condition that a starfish cover a scallop, on which DetectoRS makes a mistake. The last two rows denote the mimicry condition. In the fifth case, the stone is very similar to the scallop. Boosting R-CNN can precisely distinguish the scallop from the stone. Moreover, for the small echinus on the left-top corner, other detectors miss it for the lack of regression capacity. The proposed Boosting R-CNN accurately detects this echinus, which proves that RetinaRPN with FIoU loss can provide high-quality proposals. In the last case, although the starfish hides in the waterweeds, Boosting R-CNN still successfully detects the object.

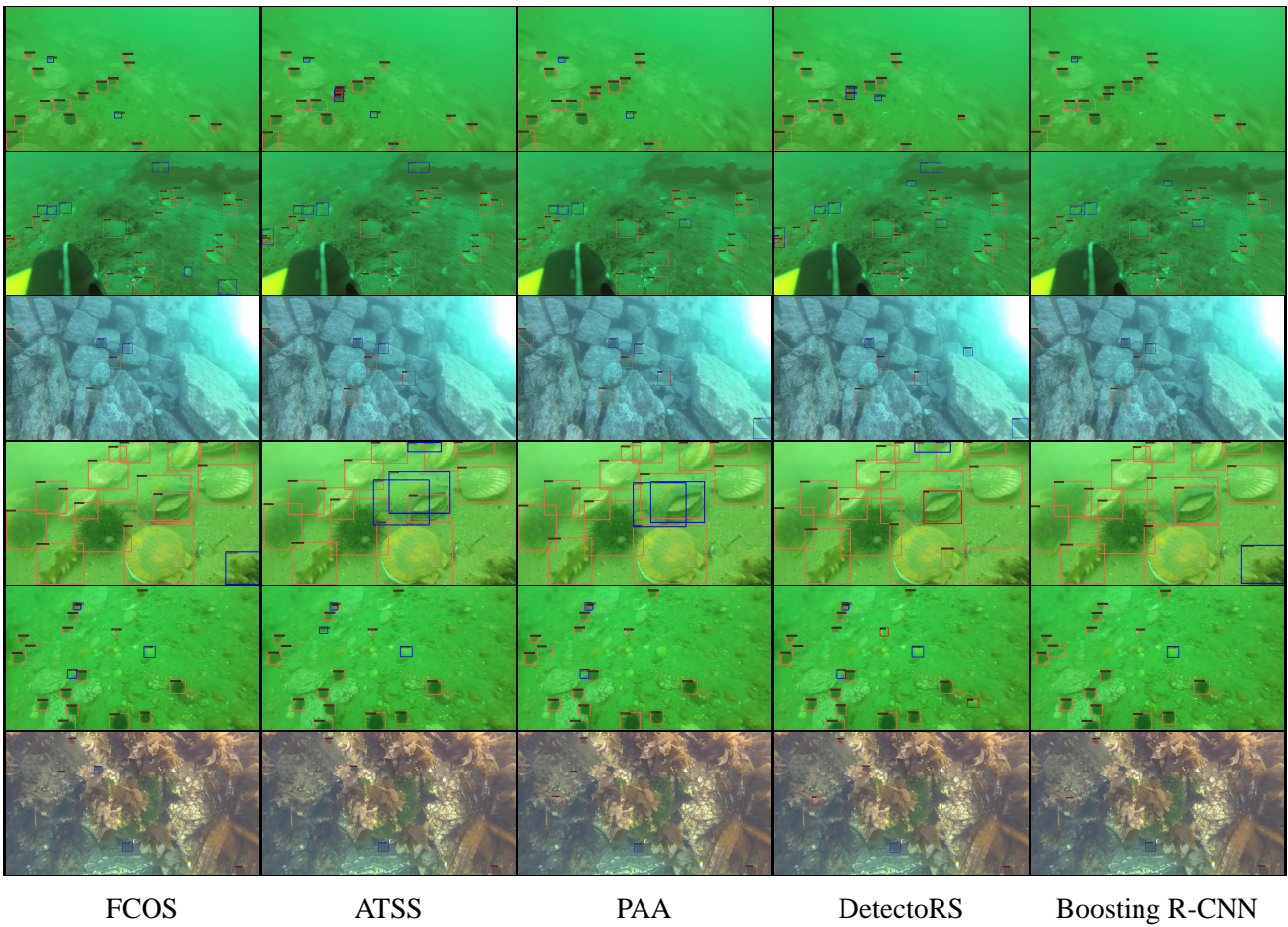

| FCOS | ATSS | PAA | DetectoRS | Boosting R-CNN |

**Figure 6:** Qualitative comparison results on the UTDAC2020 dataset. The orange boxes denote the matched predictions. The blue boxes denote the unmatched predictions. The red boxes denote the undetected ground truths. The first two rows denote the blurring and low contrast conditions. The third row denotes the unbalanced light condition. The Fourth row denotes the occlusion condition. The last two rows denote the mimicry condition.

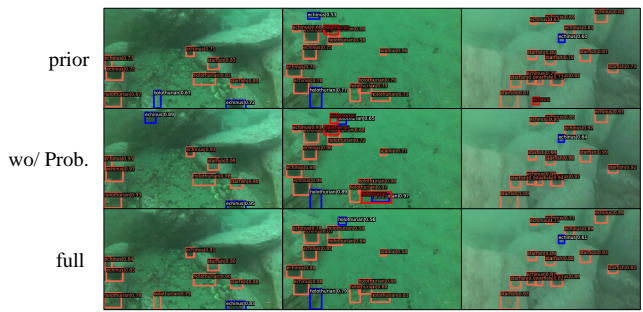

**Figure 7:** Visualization of the detection results of each variant of Boosting R-CNN. "prior" denotes replacing the second-stage classification score with the first-stage priors. "wo/ Prob." means dropping the probabilistic inference pipeline. "full" denotes the proposed Boosting R-CNN.

To further investigate the mechanism of Boosting R-CNN, we visualize the detection results of the variants.

"prior" denotes replacing the second-stage classification score with the first-stage priors, and the labels are from the classes with the highest score in the R-CNN head. "wo/ Prob." means droping the probabilistic inference pipeline and using the R-CNN head results directly. "full" denotes the proposed Boosting R-CNN. The detection scores in "full" are smaller than "wo/ Prob.", which suggests that the RetinaRPN can provide the uncertainty for the R-CNN head to avoid overfitting. For example, in the first column, the detection scores of the blue box in the top left corner ("wo/ Prob.") are decreased by the RetinaRPN. Thus, this false over-confident prediction is filtered in the final results ("full"). Besides, with the RetinaRPN and the probabilistic inference pipeline, the detection confidences are more reasonable, for Boosting R-CNN considers various uncertainty. In the second column, compared with "wo/ Prob.", no ground truth is missed in the result of Boosting R-CNN. What's more, the R-CNN trained with BR can rectify the error of the RetinaRPN. In the third column, the ground truth (the red box) in the bottom of the "prior" is missed for the RetinaRPN assigns a too small prior for the predictions. With the correction of the R-CNN, the missed ground truth can be detected by increasing the second-stage score.

## 5. Conclusion

Underwater object detection is facing new challenges such as blur, low contrast, occlusion, and mimicry conditions compared with generic object detection. In this paper, we propose a brand new two-stage underwater detector Boosting R-CNN to solve the problems mentioned above. First, the proposed RetinaRPN has a strong capacity to detect objects in blurring, low-contrast, distorted images, and provide high-quality proposals with accurate estimations of object prior probability in the occlusion condition. Second, the proposed probabilistic inference pipeline helps the detector make a prediction based on the uncertainties of the vague objects, resulting in a reasonable ranking of the prediction scores. Third, boosting reweighting is proposed to learn the second stage by the error of the first stage, which is a kind of hard example mining and helps the second stage to rectify the error at the probabilistic pipeline. The experiments on two public underwater datasets demonstrate that Boosting R-CNN outperforms other state-of-the-art detectors in underwater object detection. The competitive performances on two public generic object detection datasets show the generalization of Boosting R-CNN. Comprehensive ablation studies show the effectiveness of the proposed modules.

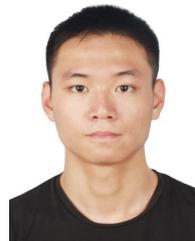

**Pinhao Song** received the B.E. degree in Mechanical Engineering in 2019, where he is currently pursuing a master's degree in computer applied technology in Peking University. His current research interests include underwater object detection, generic object detection, and domain generalization.

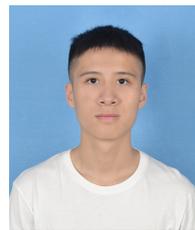

**Pengteng Li** received the B.E degree in Financial Engineering in 2020, where he is currently pursuing a master's degree in computer applied technology in Shenzhen University. His current research interests include generic object detection and reinforce learning.

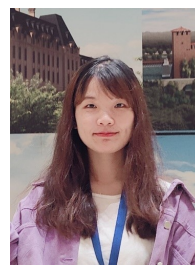

**Linhui Dai** received the B.E. degree in Information System and Information Management in 2018, and is currently working toward the Ph.D. degree with School of Electronics Engineering and Computer Science, Peking University, Beijing, China. Her current research interests include underwater object detection, open world object detection, and salient object detection.

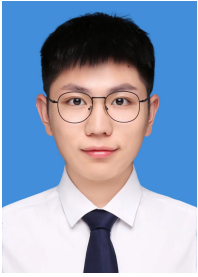

**Tao Wang** received the B.E. degree in Automation in 2020. He is currently pursuing the M.S. degree at the School of Electronic and Computer Engineering, Peking University. His research interests include person re-identification, object detection, and deep metric learning.

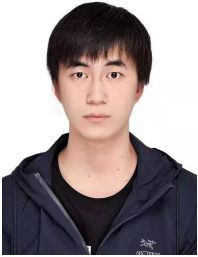

**Zhan Chen** received the B.S. degree from Hunan University(HNU), China. He is a research graduate student studying at Peking University (PKU), China. His research interest lies in machine learning and computer vision.